\documentclass{llncs}

\usepackage{makeidx}  
\usepackage{amsmath, amssymb, caption, subfig}
\usepackage{array, multirow, graphicx, siunitx}

\begin{document}
\mainmatter              
\title{Deep Learning from Label Proportions \\
for Emphysema Quantification}
\titlerunning{Deep LLP for Emphysema Quantification}  
%
\author{
Gerda Bortsova\inst{1},
Florian Dubost\inst{1},
Silas {\O}rting\inst{2},
Ioannis Katramados\inst{3},
Laurens Hogeweg\inst{3},
Laura Thomsen\inst{4},
Mathilde Wille\inst{5},
Marleen de Bruijne\inst{1,2}
}
\authorrunning{
Bortsova et al.
} 
%
\tocauthor{}
\institute{
Biomedical Imaging Group Rotterdam, Erasmus MC, The Netherlands
\and 
Department of Computer Science, University of Copenhagen, Denmark 
\and
COSMONiO, The Netherlands 
\and
Department of Respiratory Medicine, Hvidovre Hospital, Denmark
\and 
Department of Diagnostic Imaging, Bispebjerg Hospital, Denmark 
}

\maketitle              

\begin{abstract}
We propose an end-to-end deep learning method that learns to estimate emphysema extent from proportions of the diseased tissue.
These proportions were visually estimated by experts using a standard grading system, in which grades correspond to intervals (label example: 1-5\% of diseased tissue).
The proposed architecture encodes the knowledge that the labels represent a volumetric proportion.
A custom loss is designed to learn with intervals.
Thus, during training, our network learns to segment the diseased tissue such that its proportions fit the ground truth intervals.
Our architecture and loss combined improve the performance substantially (8\% ICC) compared to a more conventional regression network.
We outperform traditional lung densitometry and two recently published methods for emphysema quantification by a large margin (at least 7\% AUC and 15\% ICC), and achieve near-human-level performance.
Moreover, our method generates emphysema segmentations that predict the spatial distribution of emphysema at human level.

\keywords{emphysema quantification, weak labels, multiple instance learning, learning from label proportions}
\end{abstract}
\section{Introduction}

Estimating the volume of abnormalities is useful for evaluating disease progression and identifying patients at risk \cite{bos2014,cystic2007,Wille2016,Wille2014}.
For example, emphysema extent is useful for monitoring COPD \cite{Wille2014} and predicting lung cancer \cite{Wille2016}.

One common approach to automating the volume estimation is to segment the target abnormalities and subsequently measure their volume.
This requires expensive manual annotations, often making it infeasible to train and validate on large datasets.
Another approach is to directly regress the volume estimate (or, equivalently, a proportion of the abnormal voxels in an image).
This only needs relatively cheap weak labels (e.g. image-level visual scoring).

In this paper, we explore the weakly-labeled approach and consider it a learning from label proportions (LLP) problem \cite{Patrini2014}.
LLP is similar to multiple instance learning (MIL) in that training samples are labeled group-wise.
However, in MIL the label only signifies the presence of positive samples, whereas in LLP it is a proportion of positive samples in a group (i.e. a ``bag'').

We propose a deep LLP approach for emphysema quantification that leverages proportion labels by incorporating prior knowledge on the nature of these labels.
We consider a case where emphysema is graded region-wise using a common visual scoring system \cite{Wille2014}, in which grades correspond to intervals of the proportion of region tissue affected by emphysema.
Our method consists of a custom loss for learning from intervals and an architecture specialized for LLP.
This architecture has a hidden layer that segments emphysema, followed by a layer that computes its proportion in the input region.

Our architecture is similar to architectures proposed for MIL \cite{Wang2018}.
These methods, however, use different pooling methods and loss functions.
Very few neural-network based methods specialized for LLP were proposed \cite{dery2017weakly,li2015alter}.
\cite{dery2017weakly} learns to classify particles in high-energy physics from label proportions using a fully-connected network with one hidden layer.
\cite{li2015alter} applies LLP to ice-water classification in natural images.
This method, however, is not end-to-end: it optimizes pixel labels and network parameters in an alternating fashion.
In the case of image labeling, LLP can also be addressed more simply by using a CNN (e.g., \cite{Dubost2017,He2016}) together with a regression loss (e.g., root mean square or RMS).

Our \textbf{methodological contribution} is that we propose the first (to our knowledge) end-to-end deep learning based LLP method for image labeling.
We compare the proposed interval loss to RMS and our architecture to a conventional CNN (similar to \cite{Dubost2017,He2016}).
We perform the latter comparison in the MIL setting (when only emphysema presence labels are used for training) and in the LLP setting.
Our \textbf{application-wise contributions} are three-fold.
Firstly, we substantially outperform previous works \cite{Silas2016,Silas2018} in emphysema presence and extent prediction.
Secondly, we achieve near-human performance level in these tasks.
Thirdly, despite being trained only using emphysema proportions, our method generates emphysema segmentations that can be used to classify the spatial distribution of emphysema (paraseptal or centrilobular) at human level.

\section{Methods}

In both MIL and LLP scenarios, a dataset consists of bags of instances $\mathcal{X} = \{ x_{i} \mid i = 1...m \}$ ($m$ is a number of instances).
In MIL, each bag $\mathcal{X}$ has a binary label signifying a presence of at least one positive instance.
In LLP, this label is a proportion of positive instances in a bag.
In our case, the bag label is an ordinal variable $y \in [0, \textsf{ncat} - 1]$ (with an interpretation of emphysema grade; grade 0 corresponds to the absence of emphysema).
Values of $y$ correspond to intervals of proportion $[\textsf{thresh}_{y+1}, \textsf{thresh}_{y+2})$,
where $\textsf{thresh}$ is a vector of thresholds with the first element $\textsf{thresh}_1$ = 0 and the last element $\textsf{thresh}_{\textsf{ncat}+1}$ = 1.

\subsection{Architectures}\label{architecture}

\begin{figure}[!t]
\centering
\includegraphics[width=1.\textwidth]{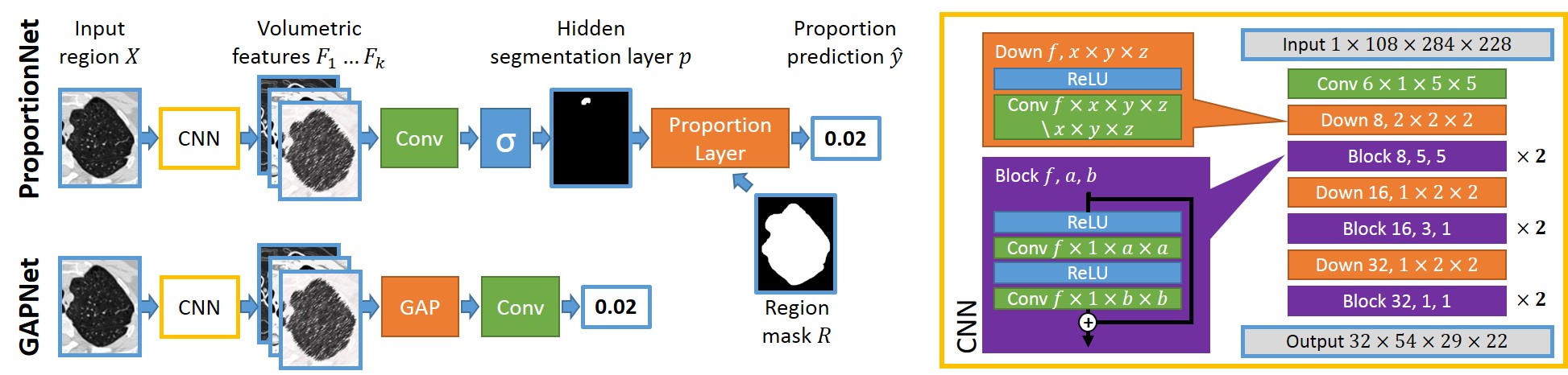}
\caption{
(a) ``Conv'': $1 \times 1 \times 1$ convolution with one output feature;
``GAP'': global average pooling;
$\sigma$: sigmoid.
(b) ``Conv'': valid convolutions with parameters \\
``\{\# of output features\}, \{kernel size\} / \{stride\}";
``Block'': residual blocks \cite{He2016}.
}
\label{arch}
\end{figure}

We call our proposed and baseline architectures ``ProportionNet'' and ``GAPNet'', respectively (see Fig. \ref{arch}).
The first layers of these architectures are the same: they both take a 3D image $X$ of a lung region as input and convert it to a set of 3D feature maps $\{F_1..F_k\}$.
The only difference between them is in how these feature maps are converted into the final output -- a proportion $\hat{y}$.

\textbf{ProportionNet} first maps the features $\{F_1..F_k\}$ to a single 3D emphysema probability map $p(X)$ and then averages the probabilities within a given region mask $R$ using ``ProportionLayer'' to obtain the emphysema proportion.
When supervised with region label proportions, ProportionNet learns to classify every instance (an image patch, in our case) in such a way that the average label in the bag (i.e. the region) is close to the ground truth proportion.

\textbf{GAPNet} first pools the feature maps $\{F_1..F_k\}$ using a global average pooling (GAP) layer (it thus aggregates instance features into bag features) and then combines these averages into the proportion prediction using a fully-connected layer.
We also consider a variation of GAPNet where GAP is replaced by masked GAP (MGAP), which averages every feature individually using $R$ as a mask.

\subsection{A Loss for Learning from Proportion Intervals (LPI)}\label{objective}

\begin{figure}%
    \centering
    {{\includegraphics[width=0.57\textwidth]{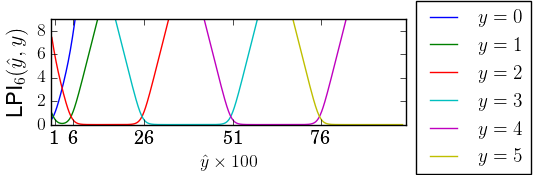} }}%
    \qquad
    {{\includegraphics[width=0.35\textwidth]{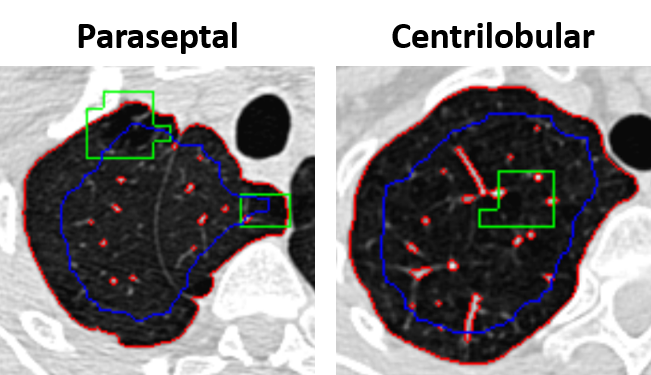} }}%
    \caption{
\textit{Left}: $\textsf{LPI}_6$ loss with all $w_c = 1$, $\alpha = 120$ and $\textsf{thresh} = (0, 0.005, 0.055, $ $0.255, 0.505, 0.755, 1)$.
\textit{Right}: images with different predominant emphysema patterns.
Green: ProportionNet segmentations; red: region mask; blue: the 10px margin for separating near-boundary detections from the rest.
    }%
    \label{loss_pat}%
\end{figure}

A good LPI loss would be near-constant when the predicted proportion $\hat{y}$ is inside the ground truth interval $[\textsf{thresh}_{y+1},
\textsf{thresh}_{y+2})$ and would increase as $\hat{y}$ goes outside the interval's boundaries. We propose a loss that approximates those properties:
$
\textsf{LPI}_{\textsf{ncat}}( \hat{y}, y ) =
\sum_{c=1}^{\textsf{ncat} - 1}{
w_c\textsf{CrossEntropy}(\sigma_\alpha(\hat{y} - \textsf{thresh}_{c+1}),
\mathbb{I}(y \geq c))
}
$,
where $\sigma_\alpha(x) = (1 + e^{-\alpha x})^{-1}$ is a sharper version of the sigmoid function, $w$ are tunable weights and $\textsf{ncat}$ is a number of categories (see Fig. \ref{loss_pat}, left). 
A $c$th term enforces that for images of grade $y \geq c$ the network predicts $\hat{y} > \textsf{thresh}_{c+1}$ and images of grade $y < c$ get $\hat{y} < \textsf{thresh}_{c+1}$.
Loss function $\textsf{LPI}_2$ that contains only the first term can be used as a MIL loss needing only binary labels ($\textsf{thresh}_2$ will be used to classify a bag into positive or negative).

In the case of ProportionNet, to the above loss we add a term enforcing the MIL assumption that in a negative bag ($y = 0$ means no emphysema) there are no positive instances:
$\textsf{MILA}(\hat{y}, y) =
\mathbb{I}(y = 0)
w_\textsf{MILA}
\big(
\frac{1}{m}\sum_i{\textsf{CrossEntropy}(p(X)_i, 0)}
\big)$.


\section{Experimental Setting}\label{exp}

\subsubsection{Dataset and Preprocessing}
Two low-dose CT scans (the baseline and follow up) were acquired from 1990 participants of the Danish Lung Cancer Screening Trial \cite{Wille2014}.
Lungs were automatically segmented and divided into 6 regions (roughly corresponding to lobes).
The image resolution was $\SI{0.78}{mm} \times \SI{0.78}{mm}$ and slice thickness was \SI{1}{mm}.
In every region, emphysema extent and the predominant pattern (paraseptal, centrilobular, panlobular) were independently identified by two observers.
The extent was assessed as a categorical grade ranging from 0 to 5 and corresponding to 0\%, 1-5\%, 6-25\%, 26-50\%, 51-75\% and 76-100\% of emphysematous tissue, respectively (as in \cite{Wille2014}).
We only used images of the right upper region and scores of one observer to train our networks (the interobserver agreement was highest in this region).
For our experiments, we randomly sampled 7 training sets of 50, 75, 100, 150, 200, 300 and 700 subjects with validation sets of the same size, except that for the largest training set the validation set contained 300 subjects.
The remaining images were assigned to the test sets.
The sampling was stratified to ensure similar inter-rater agreement in the training, validation and testing sets.

Using the region masks, we cropped images to select the target region and set all voxels outside of this region to a constant of -800 HU.
We used shifting and flipping in the axial plane to augment our training and validation sets.

\subsubsection{Network Training and Application}
All networks were trained using the Adadelta algorithm for a maximum of 150 epochs, with every epoch consisting of 600 batch iterations.
The batch size was 3.
The images were sampled in a such way that in every batch there was one healthy image (grade 0), one grade 1 image and one image of grade 2 to 5 sampled uniformly at random (meaning that e.g. grade 5 images appeared with the same frequency as grade 2).
This sampling strategy ensures that higher grade images, which are much rarer, are sufficiently represented.
For our LPI loss we used thresholds $\textsf{thresh} = (0, 0.005, 0.055, 0.165, 0.385, 0.605, 1)$, which are slightly different from the ones defined by the scoring system (given in the ``Dataset'' subsection and illustrated in Fig. \ref{loss_pat}, left).
This was because with the standard thresholds our method systematically underestimated the extent of emphysema in grade 3-5 regions, implying that these thresholds might be biased (they were not validated to correspond to real proportions).
The weights of the loss $w = (0.5, 0.1, 0.005, 0.005, 0.005)$ were chosen to prioritize accurate emphysema presence classification and account for the poorer inter-rater agreement for higher grade classification.
$w_\textsf{MILA}$ was set to 0.5 and $\alpha = 120$.

\section{Results}\label{results}

\subsubsection{Performance Metrics}

We evaluated our networks using these two metrics, averaged among the two annotators: 1) area under the receiver operating characteristic curve (AUC) measuring discrimination between grade 0 (no emphysema) and grades 1-5, and 2) average of the AUCs measuring discrimination of grades 1 vs. 2, 2 vs. 3, 3 vs. 4 and 4 vs. 5.
These metrics represent emphysema presence and extent prediction performances, respectively.
In Table \ref{results} we report means and standard deviations of these metrics computed over multiple test sets.

In Table \ref{resultsComp}, we use different metrics to be able to compare with other methods.
Intraclass correlation (ICC) was computed between  predictions of a method converted to interval midpoints and average interval midpoints of the two raters
(same as in \cite{Silas2016}).
Spearman's $r$ was computed between raw predictions and the averaged midpoints of the raters.
AUC was computed with respect to the maximum of the presence labels of the raters (as in \cite{Silas2018}).

\subsubsection{Learning from Emphysema Presence Labels (MIL)}

First, we trained GAPNet and ProportionNet for 75 epochs using $\textsf{LPI}_2$ and $\textsf{LPI}_2+\textsf{MILA}$ losses, respectively.
These losses only need binary presence labels, which makes it a MIL problem.
ProportionNet outperformed GAPNet in both presence and extent prediction by a large margin when trained on the small sets (see Table \ref{results}).
When trained on the medium and large sets, ProportionNet was similar to GAPNet in presence detection and better in extent estimation by 2-3\% of mean AUC. 

To understand the contribution of region masking to the performance of ProportionNet, we also trained MGAPNet, in which GAP was replaced by region-masked GAP, using $\textsf{LPI}_2$ (on our small sets only due to limited computational resources).
MGAPNet performed better than GAPNet in both presence (AUC $0.90\pm0.06$) and extent (AUC $0.71\pm0.03$) prediction.
ProportionNet still substantially outperformed MGAPNet.

\subsubsection{Learning from Emphysema Proportion Labels (LLP)}

We fine-tuned the GAPNet and ProportionNet previously trained in the MIL setting (see previous subsection) for another 75 epochs using $\textsf{LPI}_6$ and $\textsf{LPI}_6 + \textsf{MILA}$ losses, respectively.
ProportionNet outperformed GAPNet in both presence and extent prediction tasks in all cases, except for the medium sets, on which the presence detection performance of both networks was the same.

We also compared $\textsf{LPI}_6$ with a more conventional RMS loss.
We trained GAPNet from scratch for 150 epochs with RMS loss to regress emphysema scores (not proportions, as in this case there would be a relatively very little cost for confusing 0\% and 1-5\% grades) using the largest training set.
RMS did substantially worse than $\textsf{LPI}_6$ and worse than ProportionNet in both presence (AUC 0.94) and extent (AUC 0.72) prediction (see also Table \ref{resultsComp}).

\begin{table}[!t]
\caption{
Performance of emphysema presence detection and extent estimation (measured in average AUC over multiple test sets) of networks trained on sets of different size (in patients) and using different labels.
}
\label{results}
\begin{center}
\small
\begin{tabular}{*{6}{|c}|}
\hline
 &
\bfseries Architecture: &
\multicolumn{2}{c|}{\bfseries GAPNet} & 
\multicolumn{2}{c|}{\bfseries ProportionNet}
\\
\hline
& \bfseries Training set size\textbackslash Task: &     Presence    &      Extent     &     Presence    &      Extent \\
\hline
\parbox[t]{2mm}{\multirow{3}{*}{\rotatebox[origin=c]{90}{\bfseries MIL}}}
& small sets (50, 75, 100) &     $0.87\pm0.05$      & $0.68\pm0.06$ &  $\mathbf{0.95\pm0.01}$ & $\mathbf{0.74\pm0.02}$ \\
& medium sets (150, 200, 300) & $\mathbf{0.96\pm0.01}$ & $0.72\pm0.02$ &  $\mathbf{0.96\pm0.01}$ & $\mathbf{0.74\pm0.02}$ \\
& large set (700) &    $\mathbf{0.96}$     &     $0.76$    &    $\mathbf{0.96}$     &    $\mathbf{0.79}$ \\ 
\hline
\parbox[t]{2mm}{\multirow{3}{*}{\rotatebox[origin=c]{90}{\bfseries LLP}}}
& small sets (50, 75, 100) &     $0.90\pm0.04$      & $0.74\pm0.06$ & $\mathbf{0.94\pm0.01}$ & $\mathbf{0.79\pm0.02}$ \\
& medium sets (150, 200, 300) & $\mathbf{0.96\pm0.01}$ & $0.80\pm0.02$ & $\mathbf{0.96\pm0.01}$ & $\mathbf{0.84\pm0.01}$ \\
& large set (700) &         $0.96$         &     $0.79$    &    $\mathbf{0.97}$     &    $\mathbf{0.86}$ \\
\hline
\end{tabular} %
\end{center} %
\end{table}

\subsubsection{Comparison to Other Methods and Human Experts}

We compare our methods to two published methods, which are the most recent works that use the same dataset.
\cite{Silas2016} is an LLP method based on cluster model selection.
\cite{Silas2018} is a MIL method (trained using only presence labels) based on logistic regression.
To compare with each one of these methods, we chose a split having the same number of images or fewer for training and validation (100 and 700 subjects to compare with \cite{Silas2016} and \cite{Silas2018}, respectively).
We also evaluated several traditional densitometric methods \cite{Wille2016} and report the best result (LAA\%-950).
As can be seen from Table \ref{resultsComp}, ProportionNet and GAPNet substantially outperformed densitometry and the methods of \cite{Silas2016} and \cite{Silas2018}.

When compared with the expert raters, ProportionNet trained using the largest training set achieves ICCs of 0.84 and 0.81 between its predictions and raters' annotations, whereas the inter-rater ICC is 0.83.
It is slightly worse than the second rater in predicting the first rater's emphysema presence labels (sensitivity 0.92 vs. 0.93 when specificity is 0.9) and is as good as the first rater in predicting the second rater's labels (sensitivity 0.73, specificity 0.98).

\begin{table}[!t]
\begin{center}
\caption{
Comparison of our networks with densitometry and machine learning approaches \cite{Silas2016} and \cite{Silas2018} (they use the same dataset).
``LLP'' stands for training using extent labels and ``MIL'' -- using presence labels.
``RU'' and ``LU'' stand for right and left upper regions.
Metrics used are ICC, Spearman's $r$ and AUC.
}
\label{resultsComp}
\small
\begin{tabular}{*{11}{|c}|}
\hline
\multicolumn{1}{|c|}{\bfseries Labels:} &
\multicolumn{8}{c|}{\bfseries LLP} & 
\multicolumn{2}{c|}{\bfseries MIL}
\\
\hline
\multicolumn{1}{|c|}{\bfseries Training set size:} &
\multicolumn{4}{c|}{\bfseries 100 subjects} & 
\multicolumn{4}{c|}{\bfseries 700 subjects} &
\multicolumn{2}{c|}{\bfseries 700 subjects}
\\
\hline
\multicolumn{1}{|c|}{\bfseries Region:} &
\multicolumn{2}{c|}{\bfseries RU} & 
\multicolumn{2}{c|}{\bfseries LU} &
\multicolumn{2}{c|}{\bfseries RU} & 
\multicolumn{2}{c|}{\bfseries LU} &
\multicolumn{1}{c|}{\bfseries RU} & 
\multicolumn{1}{c|}{\bfseries LU}
\\
\hline
\bfseries Metric: & ICC & $r_s$ & ICC & $r_s$ & ICC & AUC & ICC & AUC & AUC & AUC  \\
\hline
Densitometry \cite{Wille2016} & - & 0.23 & - & 0.14 & - & 0.59 & - & 0.54 & 0.59 & 0.54 \\
\cite{Silas2016} and \cite{Silas2018} & 0.72 & - & 0.63 & - & - & - & - & - & 0.89 & 0.87 \\
GAPNet+RMS & - & - & - & - & 0.79 & 0.93 & 0.76 & 0.90 & - & - \\
GAPNet+$\textsf{LPI}_6$ & 0.77 & 0.62 & 0.74 & 0.52 & 0.82 & 0.96 & 0.76 & 0.94 & \textbf{0.96} & \textbf{0.94} \\
ProportionNet & \textbf{0.87} & \textbf{0.73} & \textbf{0.81} & \textbf{0.66} & \textbf{0.87} & \textbf{0.97} & \textbf{0.85} & \textbf{0.95} & 
\textbf{0.96} & \textbf{0.94} \\
\hline
\end{tabular} %
\end{center} %
\end{table}

\subsubsection{Emphysema Pattern Prediction}

The most common emphysema patterns are centrilobular and paraseptal (around 90\% cases in upper regions).
Paraseptal emphysema is located adjacent to lung pleura, whereas centrilobular can be anywhere in the lungs.
We designed a simple feature to discriminate between the two, given an emphysema segmentation: a ratio between the foreground volume near the boundary and inside the region (see Fig. \ref{loss_pat}).
We computed this feature using segmentations of ProportionNet trained on the largest training set.
On the test set, we obtained AUC 0.89 using the first rater (sensitivity 0.65 and specificity 0.95, same as the inter-rater ones) and AUC 0.92 using the second rater as the ground truth (sensitivity 0.61 and specificity 0.96 vs. inter-rater 0.61 and 0.91).
This performance is thus on a par with both raters.

\section{Discussion and Conclusion}\label{discussion}

We compared two architectures for MIL and LLP (ProportionNet and GAPNet) under fair conditions: the only differences were in the few final layers that aggregated instance features into bag label predictions.
ProportionNet outperformed GAPNet in both MIL and LLP settings.
We can attribute this to two factors.
Firstly, from our comparison between GAPNet and MGAPNet we learned that region masking is beneficial, probably because it acts as a location prior and makes compensating for variable region sizes unnecessary.
However, it was not the main contributor to the performance boost.
The second factor is that ProportionNet in a combination with LPI loss reflects the prior assumptions of our problem better.
When ProportionNet is trained using our MIL loss ($\textsf{LPI}_2 + \textsf{MILA}$ with $\textsf{thresh}_2 = 0.005$), the assumption is that even a very small ($>0.5\%$ volume) pathological area makes the image positive.
When trained using our LLP loss ($\textsf{LPI}_6 + \textsf{MILA}$) and proportion labels, the network is guided on approximately how much of the abnormality is in the images.
This loss also captures the interval nature of our labels better, as it allows for different predictions for same grade images.
RMS loss, for example, tries to map all examples of one grade into one value, whereas in reality same grade images often have different proportions of emphysema.
This is a probable reason for LPI outperforming RMS.

We are aware of only one work \cite{Wang2018} that performed a fair comparison of different network architectures for MIL.
In their case, a GAPNet-like network performed better than a ProportionNet-like network.
We think that to achieve a regularization effect using ProportionNet, it is crucial to select a pooling strategy and a loss that match the prior assumptions of the target problem well.

Another important advantage of ProportionNet compared to GAPNet is that it localizes the target abnormality.
In our case, the localization was good enough to classify spatial distribution of emphysema with human-level accuracy.

While in this work we focused on emphysema quantification, we expect using the proposed architecture and loss to be beneficial in other problems as well. ProportionNet can be a good regularizer for learning from visual scores related to the volume of abnormalities.
It might be a good fit for estimating the volume of intracranial calcification \cite{bos2014} and lung abnormalities \cite{cystic2007}.
Our LPI loss can be useful when labels have interval nature (e.g., \cite{cystic2007}).

\subsubsection{Acknowledgment}

This research is financed by the Netherlands Organization for Scientific Research (NWO) and COSMONiO.

\bibliography{bib}

\end{document}